\documentclass[10pt]{iopart}
\usepackage{graphicx}
\usepackage{eqnarray}
\usepackage{booktabs} 
\usepackage{tabularx}   
\usepackage{amsmath}
\usepackage{todonotes}
\usepackage{url} 
\usepackage{hyperref} 
\newcommand{\gguide}

\begin{document}

\title[Attention U-Nets for Characterizing Anomalous Diffusion in Videos]{AnomalousNet: A Hybrid Approach with Attention U-Nets and Change Point Detection for Accurate Characterization of Anomalous Diffusion in Video Data}

\author{Yusef Ahsini$^{1}$, Marc Escoto$^{2}$ and J. Alberto Conejero$^{1}$*}
\vspace{10pt}
\address{$^1$ Instituto Universitario de Matemática Pura y Aplicada (IUMPA), Universitat Politècnica de València, 46022 Valencia, Spain}
\address{$^2$ Centro de Investigación en Gestión e Ingeniería de Producción (CIGIP), Universitat Politècnica de València, 03801 Alcoi, Spain}

\ead{aconejero@upv.es}

\begin{abstract}

Anomalous diffusion occurs in a wide range of systems, including protein transport within cells, animal movement in complex habitats, pollutant dispersion in groundwater, and nanoparticle motion in synthetic materials. Accurately estimating the anomalous diffusion exponent and the diffusion coefficient from the particle trajectories is essential to distinguish between sub-diffusive, super-diffusive, or normal diffusion regimes. These estimates provide a deeper insight into the underlying dynamics of the system, facilitating the identification of particle behaviors and the detection of changes in diffusion states. However, analyzing short and noisy video data, which often yield incomplete and heterogeneous trajectories, poses a significant challenge for traditional statistical approaches.\medskip

We introduce a data-driven method that integrates particle tracking, an attention U-Net architecture, and a change-point detection algorithm to address these issues. This approach not only infers the anomalous diffusion parameters with high accuracy but also identifies temporal transitions between different states, even in the presence of noise and limited temporal resolution. Our methodology demonstrated strong performance in the 2nd Anomalous Diffusion (AnDi) Challenge benchmark within the top submissions for video tasks.

\end{abstract}

\section{Introduction}

Diffusion is a fundamental transport process in which particles spread from regions of high concentration to regions of lower concentration. 
The variance or Mean Square Displacement (MSD) of a particle measures the deviation of a particle's position $x(t)$ at time $t$ with respect to a reference position $x(0)$ over time. It is expressed as ${\rm MSD}\equiv \langle |x(t) - x_0|^2 \rangle$. Under ideal conditions, diffusion is often normal (Fickian) and characterized by a mean squared displacement (MSD) that grows linearly with time, that is, $\langle x^2(t) \rangle \propto t$. This simple relationship accurately describes various processes, including heat propagation in solids, the dispersal of small organisms, and the transport of nutrients through biological membranes \cite{klafter2005anomalous}. However, the assumption of normal diffusion frequently breaks down. Environmental complexity, medium heterogeneity, and non-equilibrium conditions often result in anomalous diffusion, i.e., when the MSD exhibits a nonlinear time dependence $\langle x^2(t) \rangle = K_{\alpha} t^{\alpha}$, with $0<\alpha\le 2$, $\alpha \neq 1$, and $K_{\alpha}$ serving as a generalized diffusion coefficient.\medskip 

The parameter $\alpha$ determines the diffusion process's time dependence and categorizes the diffusion type. When $\alpha>1$, diffusion is called \textit{superdiffusive}, while when $\alpha<1$, it is called \textit{subdiffusive}. The $K_{\alpha}$ coefficient quantifies the diffusion rate, adjusted for the anomalous scaling described by $\alpha$. Its dimensions depend on the value of $\alpha$, ensuring that the units of MSD are consistent. Anomalous behavior arises in contexts including complex animal foraging patterns \cite{vilk2022unravelling}, intracellular protein transport \cite{waigh2023heterogeneous}, and pollutant migration through porous media \cite{zhang2024modeling}. Extracting reliable estimates of $\alpha$ and $K_{\alpha}$ from short, noisy trajectories is inherently challenging, necessitating robust methods against data stochasticity and noise.\medskip

The Anomalous Diffusion Challenge (AnDi Challenge) \cite{munoz2020anomalous,munoz-gil2021objective}, first organized in 2020, sought to benchmark methods for characterizing anomalous diffusion. Participants' models were evaluated on three tasks: (i) Inference of the anomalous exponent $\alpha$, (ii) Classification of diffusion models, and (iii) Segmentation of trajectories into distinct diffusive states. The results underscored the effectiveness of machine learning (ML)-based approaches, including deep learning architectures built upon LSTM layers and convolutional blocks \cite{garibo2021efficient}, recurrent neural networks \cite{argun2021classification}, graph neural networks \cite{verdier2021learning}, WaveNet-based encoders \cite{li2021wavenet}, and carefully selected statistical features \cite{gentili2021characterization}, while providing a benchmark that has motivated subsequent studies \cite{seckler2024change, firbas2023characterization,garibo2023gramian}.\medskip

Following the first AnDi Challenge, researchers emphasized the need to refine existing approaches, mainly focusing on techniques that can detect and interpret changes in diffusive behavior. Three key challenges emerged: (i) developing methods for identifying transitions between diffusive states, (ii) determining the extent to which observed non-linear MSD dependencies reflect genuine anomalous diffusion rather than environmental heterogeneity, and (iii) discerning whether the main limitations in analysis stem from the trajectory extraction process (e.g., from videos) or the interpretation of individual trajectories themselves.\medskip

The 2nd AnDi Challenge \cite{munoz2023quantitative} aimed to address these open questions by shifting attention towards dynamic behavior and state transitions. The goal was to differentiate heterogeneity from genuine anomalous diffusion by classifying particles into one of several categories: \textit{single-state} (constant parameters over time), \textit{multiple-state} (stochastically changing states), \textit{dimerization} (transient binding events), \textit{transient-confinement} (diffusion altered by compartmentalization), and \textit{quenched-trap} (intermittent trapping events). The competition included two main tracks—one based on raw video data and another on extracted trajectories—and each track involved ensemble-level and single-trajectory tasks. In the ensemble task, participants estimated global statistics such as the number of states and the distribution of $\alpha$ and $K_{\alpha}$ values. In the single-trajectory task, participants identified internal change points (CPs) that segment the trajectory into multiple regimes, each characterized by distinct anomalous diffusion parameters and environmental constraints.\medskip 

A key difference in this new edition was incorporating videos as a primary data source, necessitating different techniques tailored to handle the temporal and spatial structure inherent in such datasets. In this iteration, participants needed to directly extract particle trajectories from fields of view (FOVs) that contained multiple interacting particles. These extracted trajectories can be represented as three-dimensional arrays of shape $(\mathrm{Particles}) \times (\mathrm{Frames}) \times (\mathrm{Dimensions})$. The complexity and heterogeneity of these data motivated our choice of a novel deep learning model architecture. 

Some methodologies were adapted from the first challenge, including combining WaveNet encoders with convolutional neural networks \cite{qu2024semantic} and models incorporating LSTM layers or standard convolutional architectures. However, the video-derived structure of the data in this second edition inspired the development of novel approaches to leverage spatiotemporal information. U-Net architectures \cite{midtvedt2021quantitative,kaestel2023deep,asghar2025u}, adapted for video processing and incorporating attention mechanisms \cite{requena2023inferring} that are effective at capturing long-range dependencies played a crucial role in enhancing the analysis and interpretation of the complex dynamics in the video data.\medskip

We employed an Attention U-Net, initially developed for biomedical image segmentation, leveraging its symmetrical encoder-decoder structure and attention mechanisms to process the three-dimensional inputs directly. The encoder portion of the U-Net captures contextual information, while the decoder provides precise localization \cite{ronneberger2015u}. Its encoder-decoder design with skip connections \cite{drozdzal2016importance} enables efficient feature extraction and reconstruction. 
Incorporating attention gates improves the model’s ability to focus on relevant features and disregard extraneous information \cite{oktay2018attention}. 
This design is well-suited for inferring states and diffusion parameters from large, complex data arrays.\medskip

\begin{figure}[htp]
    \centering    \includegraphics[width=1\linewidth]{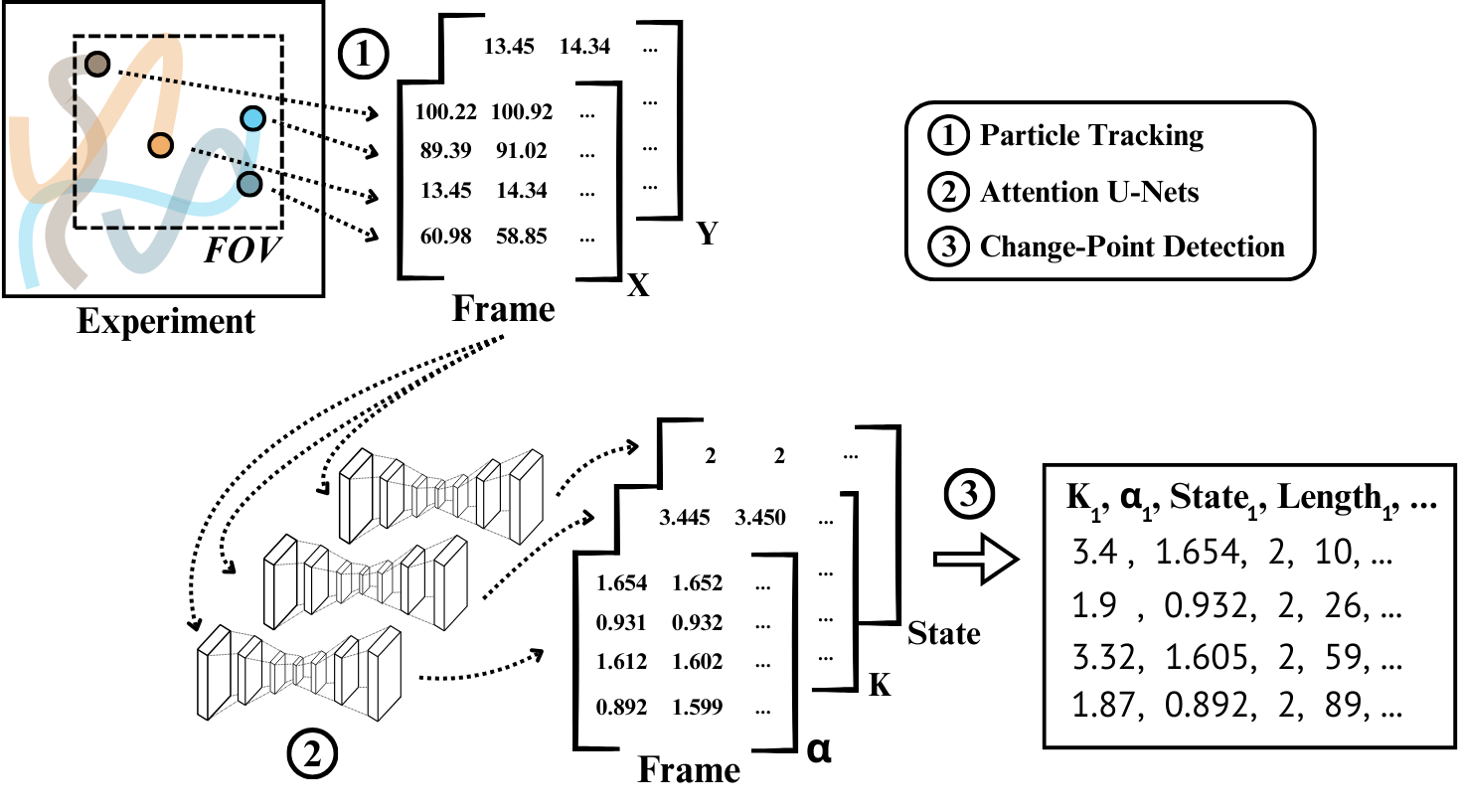}
    \caption{Diagram of the AnomalousNet Architecture }
    \label{fig:anomalnet}
\end{figure}

Building on this foundation, we developed \textit{AnomalousNet}, a comprehensive framework that integrates three core components to analyze anomalous diffusion in complex datasets. As illustrated in Figure~\ref{fig:anomalnet}, the AnomalousNet pipeline begins with particle tracking to extract spatiotemporal trajectories from FOVs, transforming raw video data into structured arrays. These trajectories are then processed by an Attention U-Net, which infers key diffusion parameters, including $K_{\alpha}$, $\alpha$, and particle states, leveraging its encoder-decoder architecture and attention mechanisms for enhanced accuracy. Finally, a change-point detection module segments trajectories into distinct regimes, identifying transitions between different diffusive behaviors. This multi-stage approach enables precise characterization of heterogeneous diffusion processes, even in noisy, high-dimensional data, and positions AnomalousNet as a robust solution for advancing the study of anomalous diffusion, see Figure \ref{fig:anomalnet}.\medskip

In the following sections, we detail our methodology, including the data simulation procedures detailed in Section \ref{sec:simulating} and in Section \ref{sec:methodology} we describe the particle tracking techniques, the Attention U-Net integration, and the CP detection algorithms. In Section \ref{sec:results}, we demonstrate that our approach not only meets the demands of the 2nd AnDi Challenge ranking within the top two submissions for the single-trajectory tasks on the video-based track and among the top six for the ensemble task in the single-trajectory track. Moreover, it also provides a generalizable framework for analyzing complex, noisy, and heterogeneous particle trajectories.
Finally, we outline the conclusions in Section \ref{sec:conclusions}.
\medskip

\section{Simulation of Experiments}
\label{sec:simulating}

The 2nd AnDi Challenge introduces a new paradigm for benchmarking methods that characterize anomalous diffusion. In particular, it focuses on understanding heterogeneous dynamics and identifying CPs in trajectories extracted from local FOVs. Each FOV represents a spatially localized snapshot of an experiment, where particles may undergo complex interactions and transitions between distinct diffusive states \cite{ernst2012fractional,manzo2015review,wang2012brownian}.\medskip

Synthetic data generation is critical since it enables controlled conditions and fully known ground truths, allowing for comprehensive validation of inference models.
We have generated a large-scale synthetic dataset to train and validate our proposed methodology using the \textit{andi-datasets} Python library \cite{munoz-gil2023objective}. This library has been specifically designed to generate data sets similar to the ones offered in the 2nd AnDi Challenge. It incorporates multiple phenomenological models that simulate anomalous diffusion under a range of conditions, including:

\begin{itemize}
    \item \textit{Single-State Model} (SSM): Particles exhibit a single anomalous diffusion state throughout their trajectory. This model applies to environments with homogeneous conditions, such as those observed for certain lipids in the plasma membrane \cite{eggeling2009direct}.
    \item \textit{Multi-State Model} (MSM): Particles stochastically switch between different diffusive states, each potentially having distinct anomalous exponents and diffusion coefficients, which accounts for the complexity of heterogeneous environments where particles can alternate between diffusion states \cite{torreno2016uncovering}.
    \item \textit{Dimerization Model} (DIM): Particle motion can be influenced by transient binding events, forming dimers that alter their mobility patterns. This phenomenon has been observed in studies of proteins in bacterial chromosome segregation \cite{tivsma2022parb} and dimerization studies in diffusion flames \cite{gleason2021pahs}.
    \item \textit{Transient-Confinement Model} (TCM): Particles experience temporary confinement within local domains, altering their effective diffusion. This behavior has been detected in studies of transient confinement zones in the plasma membrane of murine fibroblasts \cite{dietrich2002relationship} and in studies of bacterial motility within porous media \cite{bhattacharjee2019bacterial}.
    \item \textit{Quenched-Trap Model} (QTM): Particles are intermittently trapped by static energy wells, resulting in heterogeneous and intermittent dynamics, such as cytoskeleton-induced molecular pinning \cite{spillane2014high}.
\end{itemize}

Given that generating synthetic videos and utilizing them for model training is computationally intensive, we opted to decouple the processes of particle tracking and parameter inference in our model, as outlined in the introduction.
Consequently, the training data for the models consists of raw particle trajectories derived from experiment simulations. These experiments were conducted on a $512 \times 512$ pixel field, from which multiple $128 \times 128$ FOVs were extracted. This approach simulates the conditions of microscope-based single-particle tracking experiments \cite{saxton1997single,waigh2016advances}, where the field of view is limited, and the observation window is often smaller than the total sample area. Focusing on smaller FOVs minimizes boundary effects and ensures that each FOV captures a relatively homogeneous local environment.\medskip

For each simulated experiment, we generated 208 frames, although in the validation data of the 2nd AnDi-Challenge, the videos of the fields of view only consisted of 200 frames. This adjustment was made to ensure matrix compatibility with the Attention U-Net structure. The resulting trajectories were stored as three-dimensional tensors of size $64 \times 208 \times 2$, where the dimensions correspond to the maximum number of particles per FOV (64), the number of time steps (208), and the two spatial coordinates ($X$ and $Y$). Zero-padding was applied to accommodate experiments with fewer than 64 particles and to represent particles leaving the FOV. Two or more matrices were used to represent the data for cases where the FOVs contained more than 64 particles.\medskip

To ensure broad coverage of dynamic behaviors, the anomalous exponent $\alpha$ and the generalized diffusion coefficient $K_{\alpha}$ were sampled using the built-in function from the \textit{andi-datasets} package \cite{munoz-gil2023objective}, which generates random diffusion parameters extracted from a normal distribution with a mean of 1 for both $\alpha$ and $K_{\alpha}$. This approach captures subdiffusive ($\alpha < 1$), normal ($\alpha = 1$), and superdiffusive ($\alpha > 1$) regimes \cite{klafter2005anomalous,vilk2022unravelling, waigh2023heterogeneous,zhang2024modeling}.\medskip

This built-in method was modified for SSM experiments, where the $K_{\alpha}$ values were sampled from a uniform distribution within the range of $10^{-4}$ to $4$ instead of the distribution set by default. Additionally, state transition probabilities were randomly sampled using the same built-in function to ensure that different FOVs exhibit varying degrees of heterogeneity. This comprehensive sampling strategy was designed to capture various dynamic behaviors and heterogeneities across different models.
The resulting dataset includes two matrices for each FOV: one containing the raw trajectories and another with the corresponding \( \alpha \) and \( K \) values. Additionally, a third matrix encodes the state of each particle (0 = Immobile, 1 = Confined, 2 = Free, 3 = Directed), along with the label specifying the phenomenological model used in the experiment.\medskip

A database of approximately 101,250 FOVs was generated, comprising 20,250 experiments for the training set and 2,500 FOVs from 500 experiments for the validation set. The validation set is used for the inference validation. This large, diverse, and annotated database is a robust resource for training machine learning models capable of accurately inferring anomalous diffusion parameters and detecting state transitions, even in noisy and complex scenarios.\medskip

Additionally, 44 experiments with 1 FOV each (11 per phenomenological model) were created, including the raw videos and trajectory data used for generating those videos. This set is used in the system validation and it allows for the validation of the model's performance using the video tracking module and comparison with results obtained from raw trajectories.\medskip

\begin{figure}[ht]
    \centering
    \includegraphics[width=\linewidth]{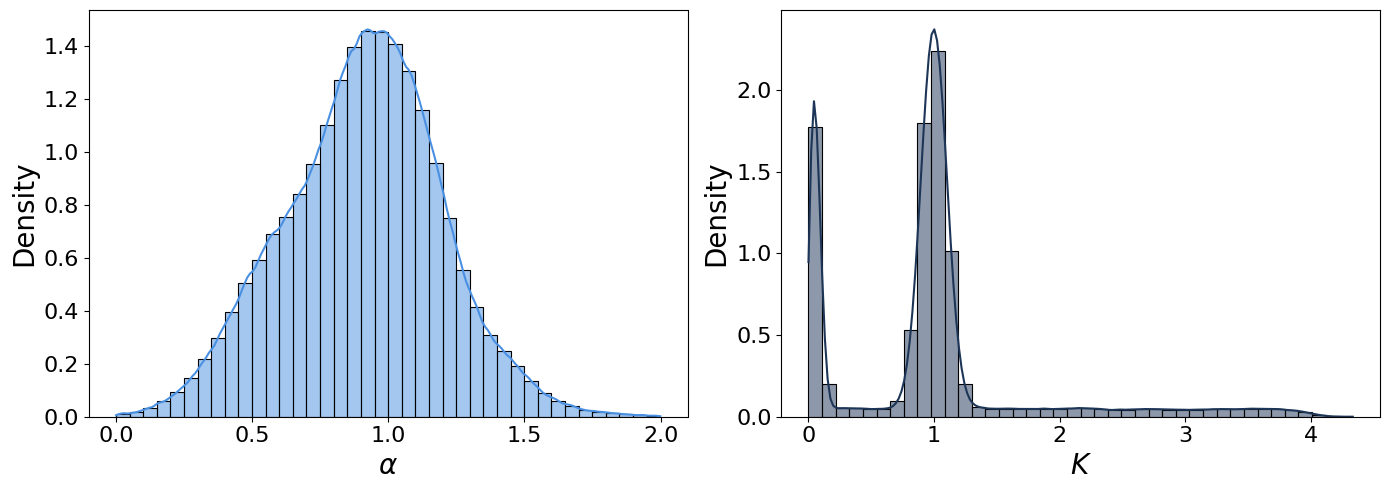}
    \caption{Density Distribution of Alpha and \(K\) Values in the Generated Data}
    \label{fig:alphas}
\end{figure}

As illustrated in Figure \ref{fig:alphas}, which displays the density distributions of $\alpha$ and $K_{\alpha}$ values in the generated data, the distribution of $\alpha$ exhibits a positively skewed profile with a peak density around $\alpha \approx 1$. The long-tail behavior towards higher values indicates the presence of asymmetric diffusion processes in the different phenomenological models, reflecting their heterogeneous environments. The distribution of $K_{\alpha}$ reveals a multimodal structure, with prominent peaks around $K_{\alpha} \approx 0$ and $K_{\alpha} \approx 1$. This behavior represents the different diffusion regimes or heterogeneous dynamics observed in the experiments. The smaller contributions beyond $K_{\alpha} > 1$ correspond to the modifications made to the generation of $K_{\alpha}$ values in the SSM experiments. This carefully designed training data distribution strategy yielded optimal results in the AnDi challenge validation set.\medskip

\section{Methodology}
\label{sec:methodology}

The proposed pipeline, termed \textit{AnomalousNet}, integrates three main steps: (i) particle tracking from raw video frames, (ii) inference of anomalous diffusion parameters and states using an Attention U-Net architecture, (iii) detection of change points to identify transitions between distinct diffusive states, and (iv) normalization of the predictions. This integrated framework is designed to address the key challenges posed by noisy observations, short trajectories, and heterogeneous dynamics often encountered in experimental single-particle tracking data.

\subsection{Particle Tracking}

Particle tracking transforms raw video sequences into structured trajectories. We have employed the TrackPy library \cite{allan2019trackpy}, which implements the Crocker-Grier algorithm \cite{crocker1996methods}, a well-established method in colloid and biophysics research. Before tracking, each $128 \times 128$ frame of the FOV was preprocessed by extending the image borders and normalizing the pixel intensities to reduce edge artifacts and improve detection stability. Particle detection was performed by choosing appropriate size thresholds and intensity cutoffs, ensuring that tracked features correspond to actual particles rather than noise or background fluctuations (diameter=3, minmass=13, separation=2.6).\medskip

After identifying particle positions in each FOV's frame, we used the command \texttt{trackpy.link} to connect these detections across consecutive frames. This process implements the Crocker-Grier algorithm, which globally minimizes the sum of squared displacements by employing a cost-minimization approach. The result is a collection of indexed trajectories, where each particle's path is defined by a sequence of $(x,y)$ positions over time.
For indexing particles in the video using a labeled frame like the VIP frame presented in the 2nd AnDi Challenge data, we first identified unique labeled regions in the VIP frame and calculated their centroid coordinates. These centroids were then matched to the detected particle positions in the initial frame by computing Euclidean distances and applying the linear sum assignment algorithm to pair each detection with the nearest label optimally \cite{2020SciPy-NMeth}.\medskip

\subsection{Model parameters' inference}

The Attention U-Net was selected for the inference step due to its ability to predict the dynamics of up to 64 particles (as defined by the model design) within a single FOV across all frames of an experiment. When a particle exits the FOV and reenters, the system interprets these events as two distinct trajectories.

This capability enables the model to capture and utilize particle correlations, encompassing their spatial positions and behavioral patterns. By modeling these interactions, the Attention U-Net learns a comprehensive representation of collective particle motion, which is then utilized for accurate predictions.\medskip

\begin{figure}[ht]
    \centering
    \includegraphics[width=\linewidth]{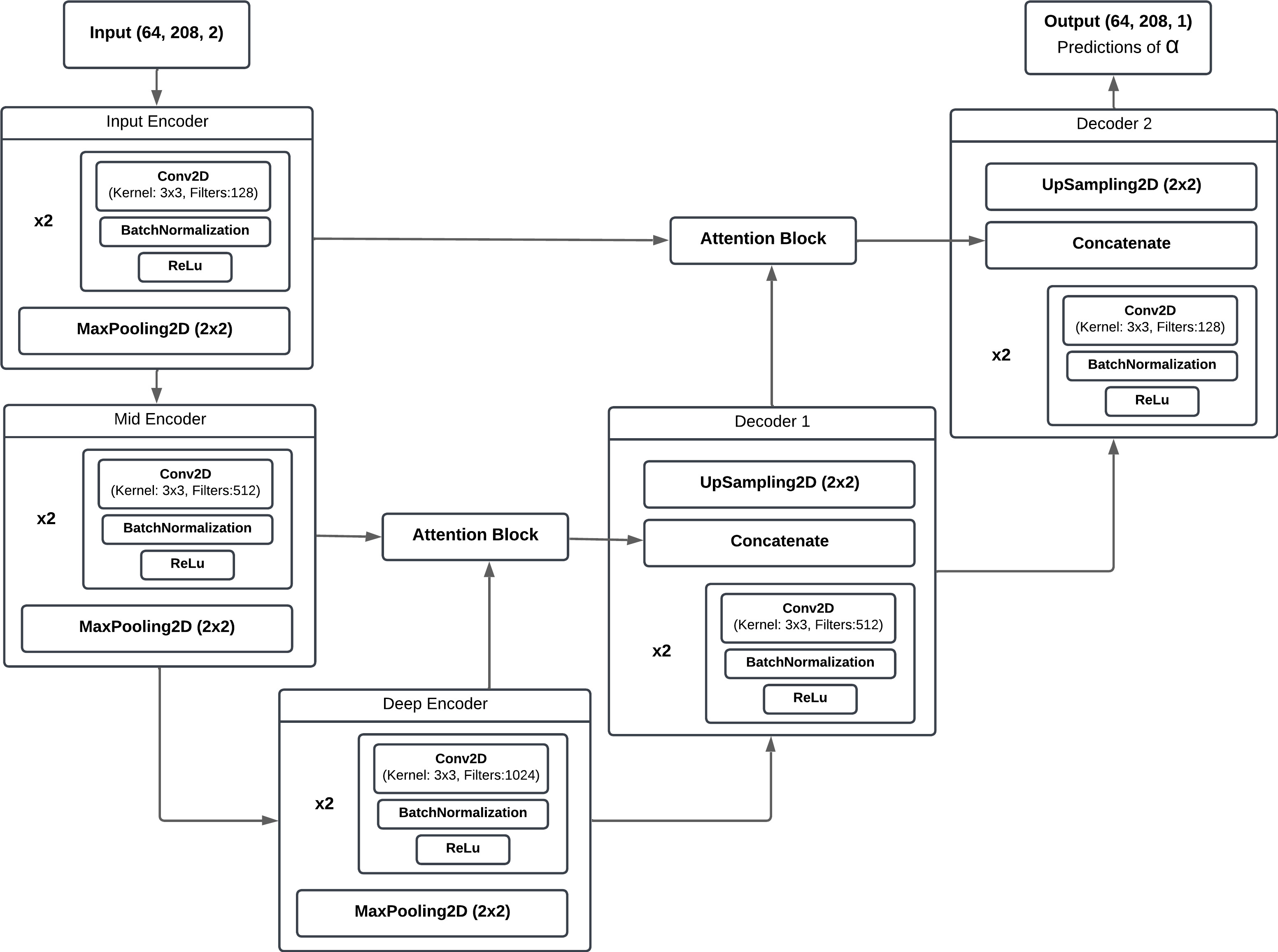}
    \caption{Attention U-Net Architecture Diagram for $\alpha$ Prediction.}
    \label{fig:model}
\end{figure}

Figure~\ref{fig:model} presents our implementation of an Attention U-Net, inspired by \cite{oktay2018attention}, designed to predict the anomalous exponent $\alpha$. This architecture enhances the standard U-Net by incorporating attention gates within the skip connections, enabling the model to focus on the most relevant features in the encoder outputs. This refinement improves its ability to capture subtle motion patterns indicative of different anomalous diffusion behaviors. The encoder progressively reduces spatial dimensions while increasing feature complexity, while the decoder mirrors this process to reconstruct the output. Attention gates ensure that only the most salient features are retained during reconstruction.\medskip

The model processes particle trajectories encoded as relative displacements between consecutive frames. As a preprocessing step, absolute coordinates $\bigl(x_i(t), y_i(t)\bigr)$ are converted into scaled relative displacements as follows:

\begin{eqnarray}
\Delta x_i(t) = \bigl(x_i(t) - x_i(t-1)\bigr) \quad
\Delta y_i(t) = \bigl(y_i(t) - y_i(t-1)\bigr).
\end{eqnarray}

Three distinct Attention U-Nets were trained, each optimized for a specific task. All models share the same general encoder-decoder architecture illustrated in Figure~\ref{fig:model}, but differ in the number of filters at each encoder and decoder level, as summarized in Table~\ref{tab:summodels}. For predicting $\alpha$, the model employs progressively deeper feature extraction layers, with 128, 512, and 1024 filter configurations at each encoder level, and utilizes the Mean Absolute Error (MAE) loss function. Similarly, the model for predicting $K_{\alpha}$ uses the same filter configurations but is trained with Keras's Mean Squared Logarithmic Error (MSLE) loss function to optimize its predictions. In contrast, the state classification model employs fewer filters (64, 96, and 128) to balance model complexity with the categorical nature of its task. Finally, we have used the \textit{SparseCategoricalCrossentropy} loss function in Keras to enhance training efficiency for the state classification model by reducing memory usage while maintaining robust gradient updates.\medskip

All models were trained using a single NVIDIA A100 GPU, with a fixed batch size 64 across all tasks. The training utilized a large dataset of 86,062 FOVs for training and 15,188 for validation. Early stopping was employed to avoid overfitting, and hyperparameters such as learning rates and filter sizes were tuned to maximize performance.\medskip

The models demonstrated strong performance across all tasks on the test dataset during the training phase. For $\alpha$, the MAE was  0.0225. For $K_{\alpha}$, the mean squared logarithmic error (MSLE) was  0.00074. State classification achieved a categorical cross-entropy loss of 0.0181. To further enhance classification accuracy, a post-processing step enforced a minimum dwell time of three consecutive frames per state, reducing transient misclassifications. These validation error metrics in the different models underestimate the true performance due to the high concentration of zeros in the matrices of the FOVs. A more comprehensive error analysis of the model is presented in the validation section.

\begin{table}[ht]
\centering
\caption{Summary of Models with Task Specifications, Encoder Levels, Filters per Level, Parameters, and Performance Metrics.}
\label{tab:summodels}
\resizebox{0.9\textwidth}{!}{
\centering
\begin{tabular}{lccccc}
    \toprule
    \textbf{Model} & \textbf{Task} & \textbf{Enc. Levels} & \textbf{Filters per Level} & \textbf{Params (M)} & \textbf{Val. Error Metric} \\
    \midrule
    \emph{Alpha\textquotesingle s} & Regression & 3 & 128--512--1024 & 28.8 & MAE: 0.0225\\
    \emph{K\textquotesingle s}       & Regression & 3 & 128--512--1024 & 28.8 & MSLE: 0.00074 \\
    \emph{State}       & Classification & 3 & 64--96--128 & 2.45 & Cross-Entropy: 0.0181 \\
    \bottomrule
\end{tabular}
}
\end{table}

\subsection{Change-Point Detection}

The CP detection step allowed us to refine the raw predictions from the Attention U-Net, consolidating them into a meaningful piecewise characterization of the underlying motion. This segmentation is crucial for differentiating genuine anomalous diffusion from heterogeneous dynamics or environmental constraints that may appear superficially similar. Inferring parameters at each frame provides rich information, but identifying CPs ultimately enables a deeper interpretation of the underlying dynamics. CP detection aims to segment each trajectory into intervals with distinct statistical properties, such as different $\alpha$ or $K_{\alpha}$ values. For CP detection, we have applied the \texttt{ruptures} Python package \cite{truong2020selective1}, a well-regarded tool in signal processing and time series analysis. In particular, we tested multiple algorithms, including Pruned Exact Linear Time (PELT), Binary Segmentation, and Bottom-Up segmentation, combined with various regularization cost models to prevent overfitting (e.g., L1, L2, and linear). Ultimately, an L2 cost model with a sliding-window approach provided the best detection sensitivity and computational efficiency balance. By examining short overlapping segments of the parameter time series, we could identify abrupt shifts in the mean level of $\alpha$ or $K_{\alpha}$ or changes in state distributions.\medskip

\subsection{Predictions Normalization}

Once the Attention U-Nets have produced frame-by-frame predictions of the parameters \(\alpha(t)\), \(k(t)\), and a discrete state label for each time point, we refine these preliminary estimates through a brief smoothing step before segmentation. This refinement targets isolated, short-lived reversals in the predicted state, which can arise from minor inaccuracies or measurement noise. Specifically, any one- or two-frame excursions to a contrasting label are reclassified to the nearest persistent state, thus ensuring that rapid, spurious transitions do not overshadow the underlying dynamics. This smoothing stabilizes the discrete label assignments and facilitates a more precise delineation of genuine behavioral shifts once change-point detection is applied.\medskip

The subsequent segmentation hinges on identifying temporal intervals where the parameters can be considered relatively uniform. The trajectory is construed as a single-state process if no such transition points are detected. In that case, the median computed across all frames determines the final assignment for \(\alpha\) and \(k\). At the same time, the discrete state is simply taken to be the most frequent label among those predicted initially. Conversely, suppose one or more breakpoints are uncovered. In that case, the trajectory is subdivided into multiple intervals, each characterized by its own piecewise-constant values of \(\alpha\), \(k\), and a discrete state label.\medskip

Following establishing these refined parameters at the single-particle level, an ensemble-level characterization is achieved by aggregating the segment-specific parameters across all particles within a field of view (FOV). Each segment, originating from a single-state or multi-state trajectory, contributes its representative \(\alpha\) and \(k\) values to a collective pool. Clustering techniques are employed to discern the overarching dynamical regimes present in an ensemble, and this aggregated data is analyzed. For instance, a two-cluster K-means algorithm is utilized when multiple states are detected to identify distinct groups within the \(\alpha\)-\(k\) parameter. The statistical properties of each identified cluster, including mean values and standard deviations of \(\alpha\) and \(k\), are then computed.\medskip

\section{Validation results}
\label{sec:results}

We evaluate our AnomalousNet system based on the metrics defined in the 2nd AnDi Challenge \cite{munoz2023quantitative}. This evaluation is divided into three parts:
\begin{enumerate}
    \item \textbf{Inference Validation:} The performance of the inference module is assessed at the frame level using the challenge-specific metrics.
    \item \textbf{System Validation:} In the system validation, we validate the models using the AnDi Challenge metrics and methodology, applying both ensemble-level and single-trajectory tasks for videos (Track 1) and raw trajectories (Track 2). 
    \item \textbf{2nd AnDi Challenge Benchmark:} The complete system is evaluated using the 2nd AnDi Challenge benchmark.
\end{enumerate}

Below we summarize the error metrics and methodology employed during the challenge. We refer the reader to \cite{munoz2023quantitative} for further details. Equations \eqref{eq:cp_distance}--\eqref{eq:F1_score} describe the error metric for the Single-Trajectory tasks, while Equation \eqref{eq:Wasserstein} pertains to the Ensemble task.

\begin{enumerate}
\item \textit{Change Point Detection (CP)}: The CP metric compares ground-truth CPs with predicted CPs. For a ground-truth CP at time \(t^{(\mathrm{GT})}_i\) and a predicted CP at time \(t^{(\mathrm{P})}_j\), we define the gated absolute distance as
\begin{eqnarray}
d_{i,j} & = & \min\Bigl(|t^{(\mathrm{GT})}_i - t^{(\mathrm{P})}_j|,\, \varepsilon_{\mathrm{CP}}\Bigr),
\label{eq:cp_distance}
\end{eqnarray}
with \(\varepsilon_{\mathrm{CP}} = 10\). An optimal pairing between ground-truth and predicted CPs is determined via the Hungarian algorithm, which minimizes the total distance. A pair is considered a true positive (TP) if \(d_{i,j} < \varepsilon_{\mathrm{CP}}\) and unmatched or poorly matched if CPs are counted as false positives (FP) or false negatives (FN). Then, the \textit{Jaccard similarity coefficient} (JSC) is computed as
\begin{eqnarray}
\mathrm{JSC} & = & \frac{\mathrm{TP}}{\mathrm{TP} + \mathrm{FP} + \mathrm{FN}},
\label{eq:jsc}
\end{eqnarray}
and the root mean square error (RMSE) over paired CPs is given by
\begin{eqnarray}
\mathrm{RMSE} & = & \sqrt{\frac{1}{N}\sum_{i,j\,:\, d_{i,j} < \varepsilon_{\mathrm{CP}}} \Bigl(t^{(\mathrm{GT})}_i - t^{(\mathrm{P})}_j\Bigr)^2},
\label{eq:rmse}
\end{eqnarray}
where \(N\) is the number of paired CPs with \(d_{i,j} < \varepsilon_{\mathrm{CP}}\).

\item \textit{Parameter Estimation}: We segment the trajectories based on the detected CPs. We characterize each segment by an anomalous diffusion exponent \(\alpha\) and a generalized diffusion coefficient \(K\). Then, the mean absolute error (MAE) for \(\alpha\) is defined as
\begin{eqnarray}
\mathrm{MAE}(\alpha) & = & \frac{1}{N}\sum_{i=1}^{N} \Bigl|\alpha^{(\mathrm{GT})}_i - \alpha^{(\mathrm{P})}_i\Bigr|,
\label{eq:mae_alpha}
\end{eqnarray}
and the mean squared logarithmic error (MSLE) for \(K\) is given by
\begin{eqnarray}
\mathrm{MSLE}(K) & = & \frac{1}{N}\sum_{i=1}^{N} \Bigl[\ln\bigl(K^{(\mathrm{GT})}_i + 1\bigr) - \ln\bigl(K^{(\mathrm{P})}_i + 1\bigr)\Bigr]^2.
\label{eq:msle_k}
\end{eqnarray}

\item \textit{Diffusion State Classification}: We evaluate the diffusion state classification using the F1-score:
\begin{eqnarray}
F1 & = & \frac{2\,\mathrm{TP}_c}{2\,\mathrm{TP}_c + \mathrm{FP}_c + \mathrm{FN}_c},
\label{eq:F1_score}
\end{eqnarray}
where \(\mathrm{TP}_c\), \(\mathrm{FP}_c\), and \(\mathrm{FN}_c\) denote the counts for correctly classified states, false positives, and false negatives, respectively.

\item \textit{Ensemble Task}: In the ensemble task, we compare the predicted distributions \(P(K)\) and \(P(\alpha)\) with the ground-truth distributions \(Q(K)\) and \(Q(\alpha)\) using the first Wasserstein distance (W1):
\begin{eqnarray}
W_1(P,Q) & = & \int_{\mathrm{supp}(Q)} \Bigl| \mathrm{CDF}_P(x) - \mathrm{CDF}_Q(x)\Bigr| \, dx,
\label{eq:Wasserstein}
\end{eqnarray}
where \(\mathrm{CDF}_P\) and \(\mathrm{CDF}_Q\) denote the cumulative distribution functions of \(P\) and \(Q\), respectively.
\end{enumerate}

\subsection{Inference validation results}

The local validation phase assesses the accuracy and robustness of the Attention U-Net module within the \textit{AnomalousNet} pipeline for inferring anomalous diffusion parameters ($\alpha$ and $K$) and classifying motion states from raw trajectories generated to mimic diverse physical models. As it is aforementioned indicated, the evaluation is conducted on a subset of $2,500$ FOVs, excluding change point detection. It focuses specifically on the model's performance of inferring at each frame of particle movement.\medskip

\begin{figure}[ht]
\begin{center}
    \includegraphics[width=\linewidth]{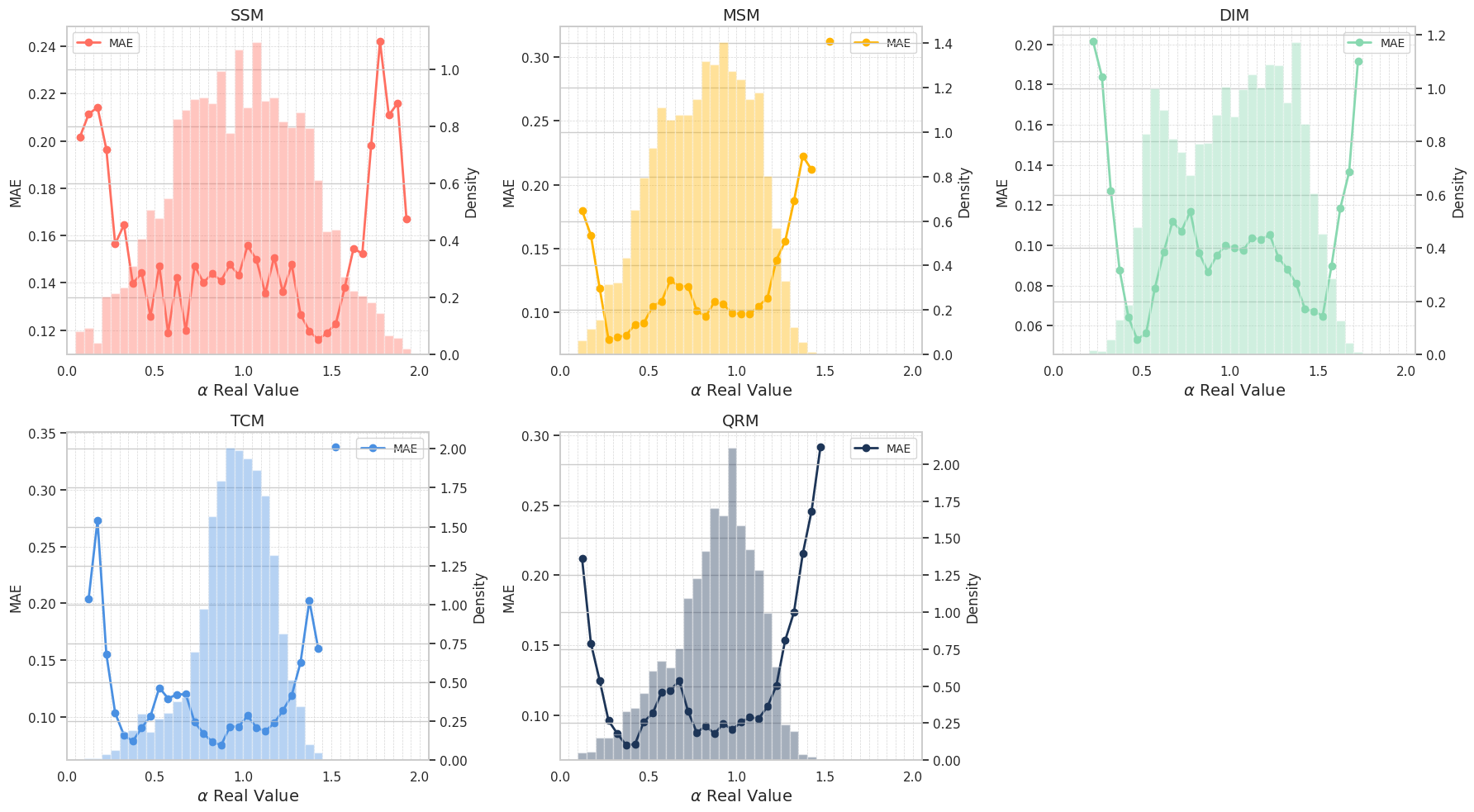}
    \caption{Histogram of the anomalous diffusion exponent $\alpha$ across different models in the inference validation dataset across different physical models. The line represents the MAE of the predictions for each value of $\alpha$.}
    \label{fig:mae_alphas}
\end{center}
\end{figure}

Figure~\ref{fig:mae_alphas} presents the MAE for $\alpha$ predictions across various physical models. The results indicate the model achieves consistent and accurate predictions in single-state and multi-state scenarios. The MAE is the lowest for systems with uniform parameter distributions, such as SSM, where particles exhibit constant dynamics throughout their trajectories. In contrast, we observe slightly higher errors in more heterogeneous models like the QTM due to the complex intermittent trapping dynamics, which introduce abrupt changes in motion that are challenging to capture.\medskip

\begin{figure}[ht]
    \centering
    \includegraphics[width=\linewidth]{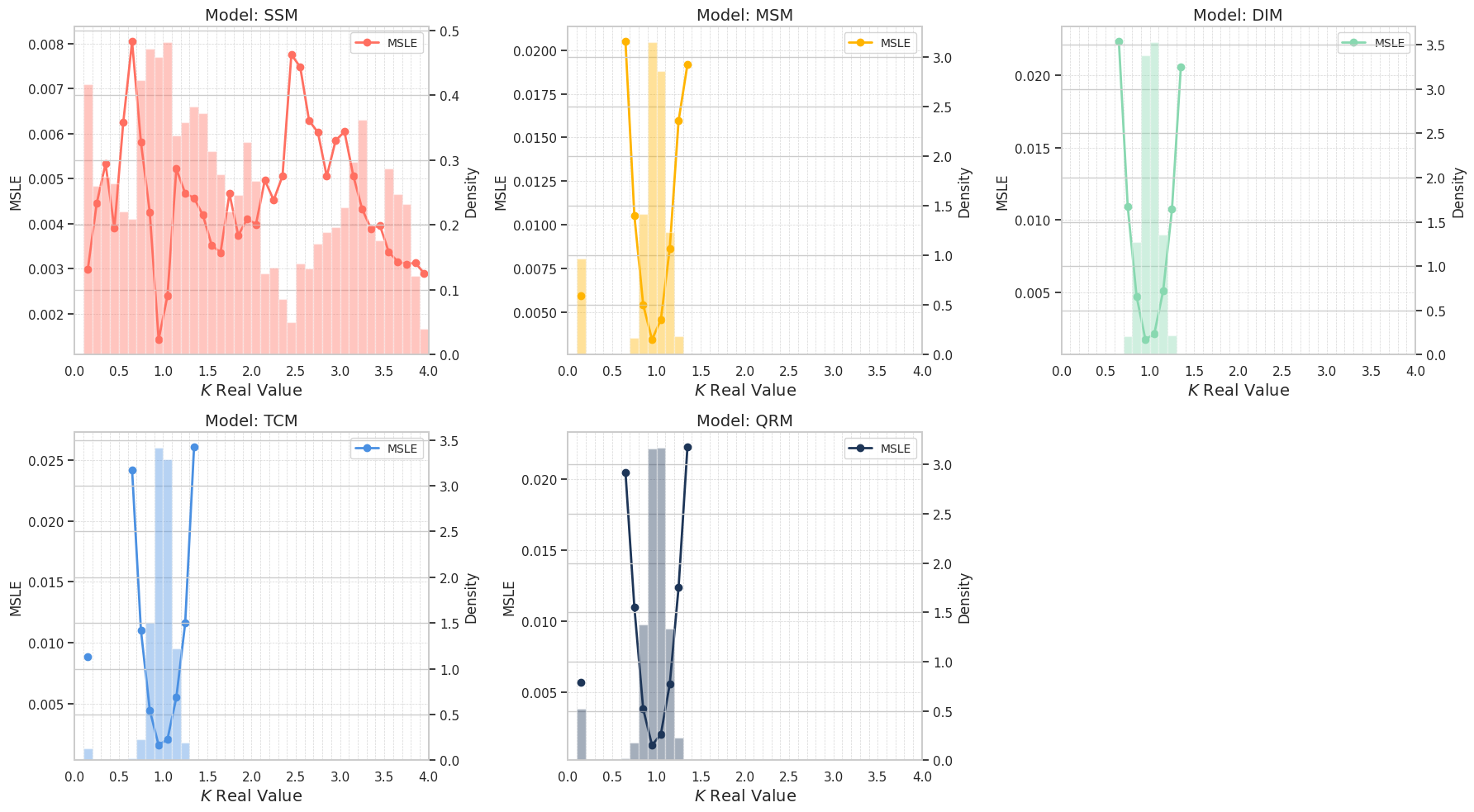}
    \caption{Histogram of the anomalous diffusion coefficient $K$ across different models in the inference validation dataset across different physical models. The line represents the MSLE of the predictions for each value of $K$.}
    \label{fig:msle}
\end{figure}

In the same line, Figure~\ref{fig:msle} illustrates the MSLE for $K's$ predictions across different models. This logarithmic metric is particularly well-suited for analyzing $K$ due to its wide range of possible values, spanning several orders of magnitude. The results show that the pipeline robustly handles both low and high $K$ regimes, with errors remaining within acceptable thresholds across all models. The SSM offers the best validation results across all phenomenological models, as expected due to the lack of state changes in the particle motion. At the same time, errors are marginally higher in TCM and QTM scenarios, where rapid fluctuations in particle mobility complicate the inference process. \medskip

\begin{figure}[htp]
    \centering
    \includegraphics[width=\linewidth]{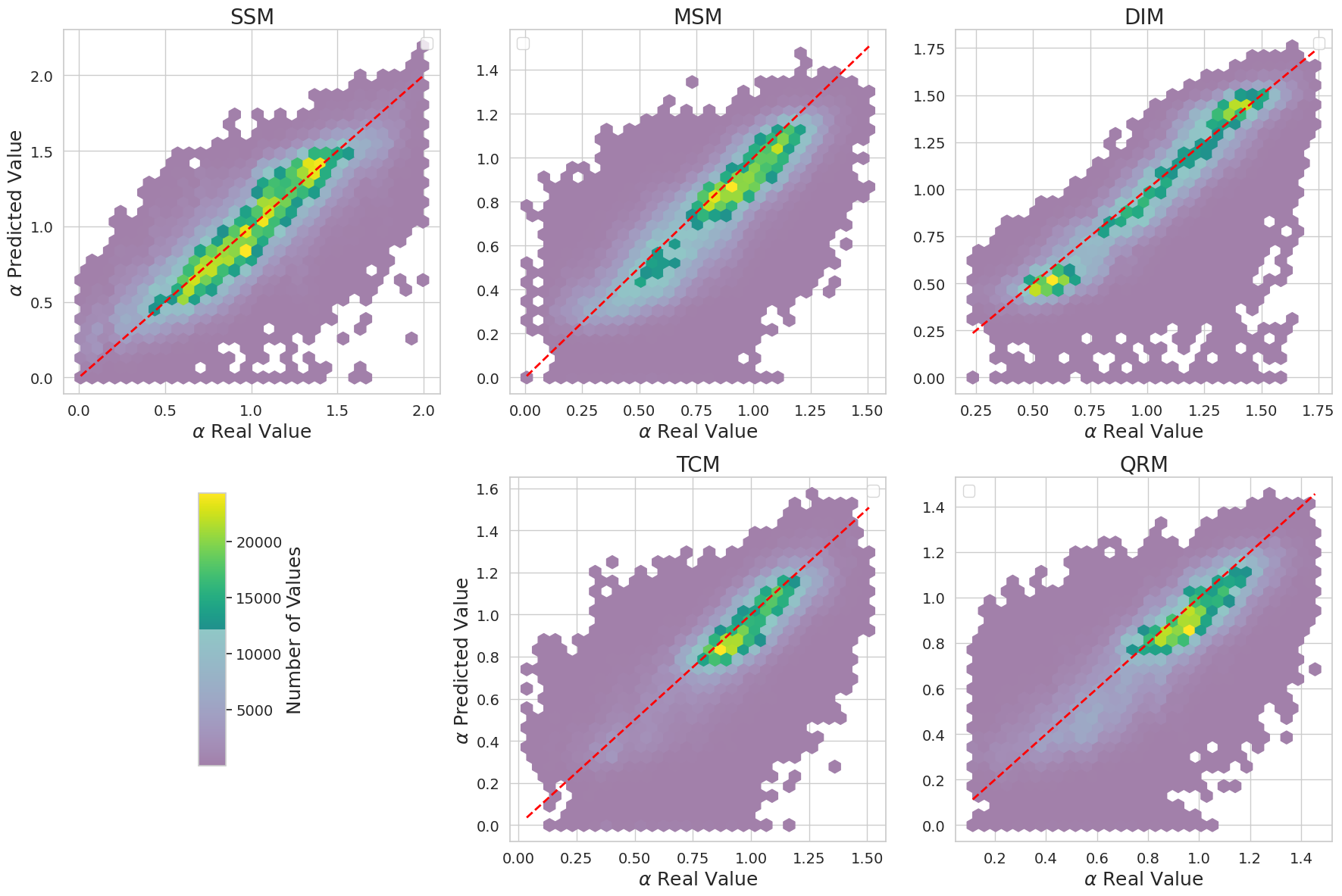}
    \caption{Heatmap of regression accuracy for predictions of the anomalous diffusion exponent $\alpha$ across different physical models.}
    \label{fig:heatmap_alpha}
\end{figure}

We also present an alternative representation of errors in the parameters inference with heatmaps. In Figure~\ref{fig:heatmap_alpha}, the regression accuracy for $\alpha$ shows a near-perfect diagonal alignment, indicating a strong agreement between predictions and ground truth values. Slight deviations are observed at extreme $\alpha$ values (e.g., $\alpha \approx 0.2$ or $\alpha \approx 2.0$), corresponding to subdiffusive and superdiffusive regimes, respectively. These deviations are likely due to limited training data in these regions and the inherent noise in trajectories exhibiting such dynamics.\medskip

\begin{figure}[htp]
    \centering
    \includegraphics[width=\linewidth]{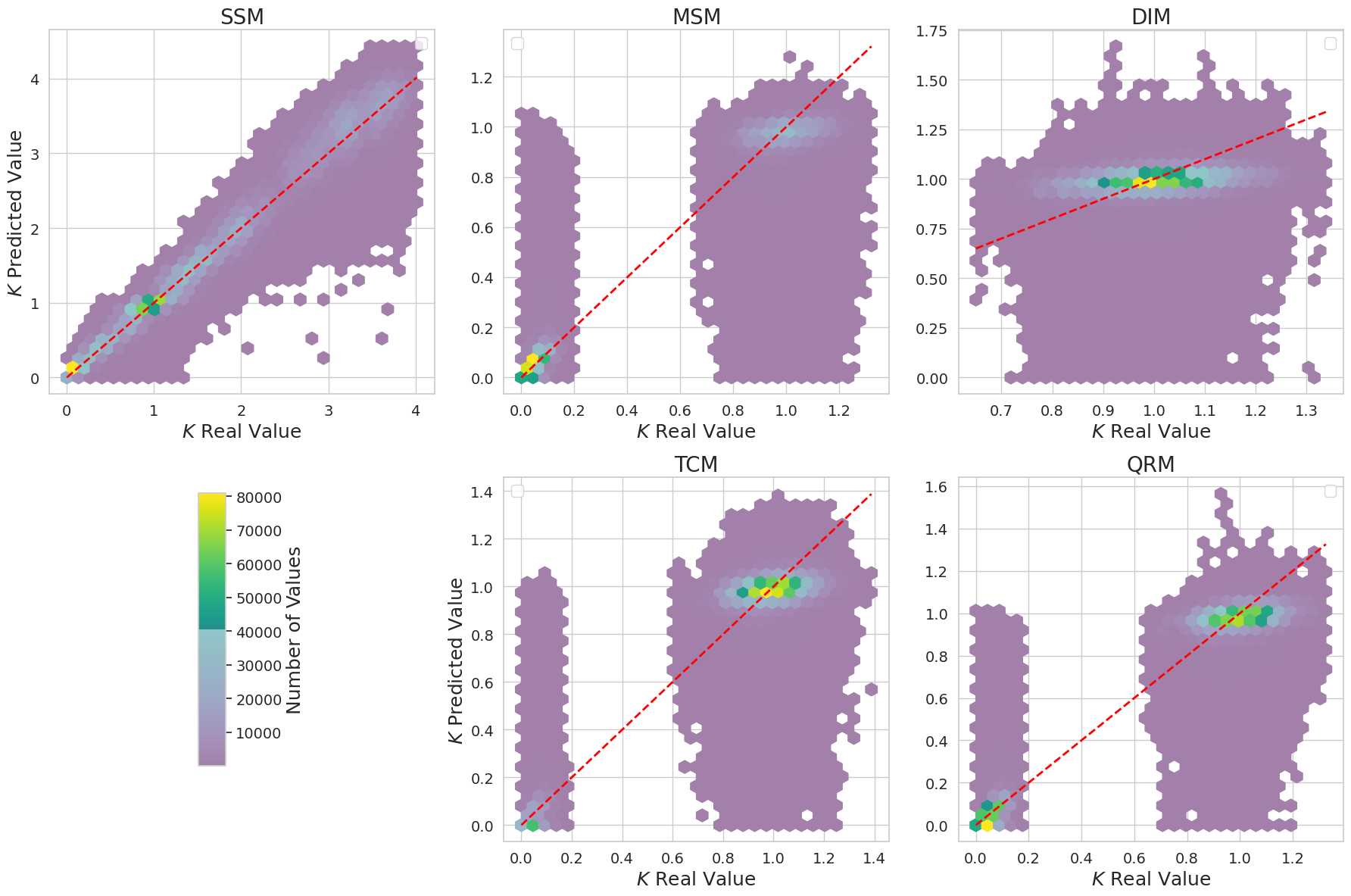}
    \caption{Heatmap of regression accuracy for predictions of the anomalous diffusion coefficient $K$ across different physical models.}
    \label{fig:heatmap_k}
\end{figure}

Figure~\ref{fig:heatmap_k} highlights the regression accuracy for $K$. Predictions exhibit high fidelity across the mid-range values of $K$, with minor discrepancies observed for extreme values (e.g., $K \ll 10^{-4}$ or $K \gg 10^{1}$). These discrepancies stem from the sparse representation of such values in the training dataset and the increased sensitivity of $K$ to noise in particle trajectories.\medskip

\begin{figure}[ht]
    \centering
    \includegraphics[width=0.75\linewidth]{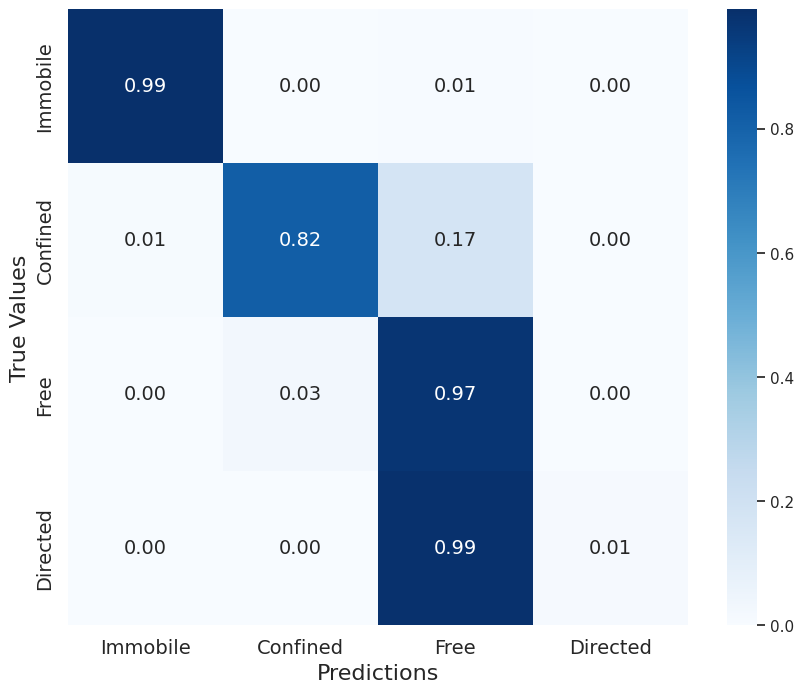}
    \caption{Normalized confusion matrix illustrating classification performance across motion states.}
    \label{fig:confusion_matrix}
\end{figure}

Finally, Figure~\ref{fig:confusion_matrix} presents the normalized confusion matrix for classifying immobile, confined, free, and directed states. The model distinguishes immobile and free‐diffusion states, achieving better than 97\% accuracy. Confined motion exhibits a slightly higher misclassification rate (82\% accuracy, with a 17\% confusion rate against free diffusion), reflecting the subtlety of slow vs.\ moderate dynamics in certain intermediate regimes. The directed state, sparsely represented in the training set (associated with $\alpha>1.9$), suffers from low recall, reaching only about 1\% accuracy due to confusion with free diffusion.\medskip

\subsection{System validation results}

In this subsection, we validate AnomalousNet’s performance using a validation set of 44 experiments on raw videos and the trajectory data used for generating those videos.
This addition enables us to compare AnomalousNet’s predictions on videos and on raw trajectories. In accordance with the challenge guidelines, predictions from video data were made on only a subset of the particles—called VIP particles (15 VIPs per field of view in our data).\medskip

\begin{figure}[htp]
    \centering
    \includegraphics[width=0.70
    \linewidth]{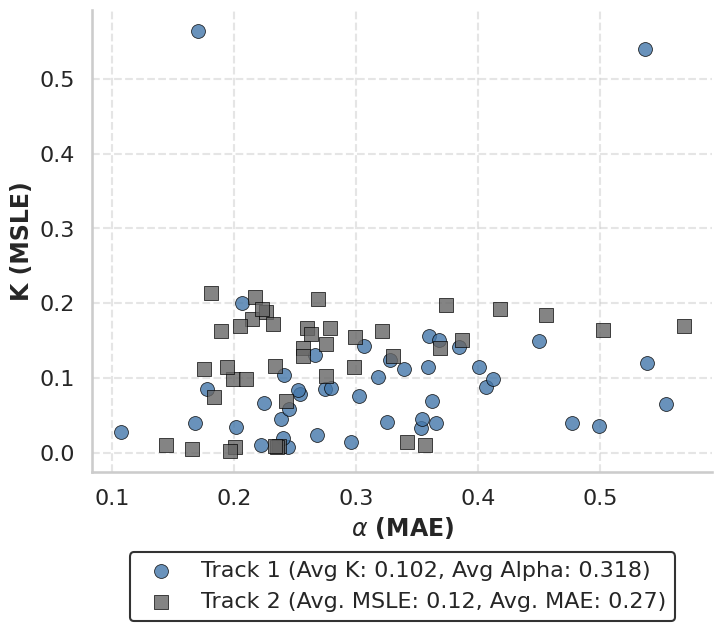}
    \caption{Comparison of AnomalousNet’s prediction errors For both Track 1 (blue circles) and Track 2 (gray squares). The scatter plot shows the MAE of the anomalous diffusion exponent \(\alpha\) (horizontal axis) versus the MSLE of the generalized diffusion coefficient \(K\) (vertical axis).}
    \label{fig:xy}
\end{figure}

Figure \ref{fig:xy} compares the MAE for \(\alpha\) and the MSLE for \(K\) for both tracks.  AnomalousNet exhibits overall comparable performance across both tracks, showing a slightly lower MAE for $\alpha$ predictions in Track 1 and providing marginally more accurate \(K\) predictions based on MSLE for Track 2.\medskip

The results in Table \ref{tab:metrics} illustrate the pipeline's performance differences between raw trajectories and videos in our system for inferring anomalous diffusion parameters. Notably, the video modality achieves a dramatically lower RMSE for change point detection (0.09 vs. 1.38), indicating a closer match between the predicted and actual change points than raw trajectories. Similarly, the Jaccard Similarity Coefficient for change points (JSC CP) is higher for video (0.54 vs. 0.46), reflecting improved overlap between the predicted and true change points—likely due to the lower number of detected change points in the video data.\medskip

\begin{table}[h!]
\centering
\resizebox{\textwidth}{!}{%
\begin{tabular}{lccccccc}
\toprule
Track & RMSE CP & JSC CP & $\alpha$ (MAE) & \(K\) (MSLE) & State (F1) & $\alpha$ Ensemble (W1) & \(K\) Ensemble (W1) \\
\midrule
Track 1 (Video)        & 0.09 & 0.54 & 0.32 & 0.10 & 0.93 & 0.26 & 0.41 \\
Track 2 (Raw trajectories) & 1.38 & 0.46 & 0.27 & 0.12 & 0.87 & 0.14 & 0.10 \\
\bottomrule
\end{tabular}%
}
\caption{Comparison of metrics performance on the system validation dataset.}
\label{tab:metrics}
\end{table}

Regarding parameter estimation, the MAE for the anomalous diffusion exponent $\alpha$ is slightly higher for video data (0.32 vs. 0.27). In contrast, the MSLE for the diffusion coefficient (\(K\)) is marginally lower (0.10 vs. 0.12), demonstrating comparable performance between the two modalities. Additionally, the state classification performance, measured by the F1 score, is marginally better for video data (0.93 vs. 0.87), though both approaches yield high accuracy.\medskip

In the ensemble task, where the objective is to predict the density distributions of $\alpha$ and \(K\) evaluated using the First Wasserstein distance (W1), both the $\alpha$ Ensemble (W1) and \(K\) Ensemble (W1) scores are significantly lower when using raw trajectories compared to video data. This result indicates that the raw trajectory-based predictions yield more robust estimates of the underlying density functions, which is not unexpected given that the Attention U-Net is trained on raw trajectories.\medskip

In summary, both video and raw trajectory approaches perform similarly in the single trajectory inference task, delivering comparable results in parameter estimation and state classification. This outcome validates the methodology of combining a particle tracking module with an inference module trained in raw trajectories. However, the ensemble task reveals that noise in the trajectories extracted by the particle tracking module leads to less robust predictions of the density functions.\medskip

\subsection{2nd AnDi Challenge Benchmark}

Finally, we present the performance of AnomalousNet in the 2nd ANDI Challenge Benchmark \cite{munoz_gil_2024_14281479}, providing a clear and easily comparable evaluation of the system. The results reported here differ from those presented for AnomalousNet in the Challenge due to modifications made to the training data and inference code during the development of this article. The parameters of the dataset are available in the supplementary material of \cite{\cite{munoz2023quantitative}}. \medskip

Table \ref{tab:results} and \ref{tab:results2} display the results of AnomalousNet on the benchmark data for Tracks 1 and 2. In these tables, the error metric corresponding to the ensemble task (ens.) is computed as the arithmetic mean. In contrast, the remaining metrics are calculated as a weighted average based on the number of trajectories, following the methodology used in the competition. It can be observed that a robust performance is obtained in both tracks compared to the competition results in most metrics. The only weak point of the system showcased in this validation is the low performance of the change-point detection module.\medskip

\begin{table}[htp]
\centering
\caption{Results for Track 1 (Video Track) for the 2nd AnDi Benchmark.}
\label{tab:results}
\begin{tabularx}{\textwidth}{l *{8}{>{\centering\arraybackslash}X}}
\toprule
\textbf{Exp} & \textbf{Num.\ trajs} & \textbf{RMSE CP} & \textbf{JSC CP} & \textbf{$\alpha$ (MAE)} & \textbf{$K$ (MSLE)} & \textbf{state (F1)} & \textbf{$\alpha$ ens. (W1)} & \textbf{$K$ ens. (W1)} \\
\midrule
1 & 300 & 0   & 0     & 0.362 & 0.110 & 1   & 0.285 & 0.181 \\
2 & 300 & 0   & 0.017 & 0.389 & 0.134 & 1   & 0.198 & 0.967 \\
3 & 300 & 3.000 & 0.459 & 0.242 & 0.073 & 0.879 & 0.300 & 0.085 \\
4 & 300 & 3.317 & 0.246 & 0.204 & 0.074 & 0.654 & 0.347 & 0.071 \\
5 & 300 & 5.020 & 0.076 & 0.198 & 0.069 & 0.997 & 0.211 & 0.186 \\
6 & 300 & 0   & 0.485 & 0.357 & 0.087 & 1   & 0.244 & 0.326 \\
7 & 300 & 5.099 & 0.013 & 0.361 & 0.105 & 1   & 0.177 & 0.104 \\
8 & 300 & 0   & 0.849 & 0.447 & 1.188 & 0.970 & 0.246 & 0.228 \\
9 & 300 & 3.518 & 0.009 & 0.849 & 0.151 & 0.211 & 0.294 & 0.111 \\
\midrule
\textbf{Avg.} & \textbf{300} & \textbf{2.217} & \textbf{0.239} & \textbf{0.379} & \textbf{0.221} & \textbf{0.857} & \textbf{0.256} & \textbf{0.251} \\
\bottomrule
\end{tabularx}
\end{table}

\begin{table}[htp]
\centering
\caption{Results for Track 2 (Trajectories Track) for the 2nd AnDi Benchmark.}
\label{tab:results2}
\begin{tabularx}{\textwidth}{l *{8}{>{\centering\arraybackslash}X}}
\toprule
\textbf{Exp} & \textbf{Num.\ trajs} & \textbf{RMSE CP} & \textbf{JSC CP} & \textbf{$\alpha$ (MAE)} & \textbf{$K$ (MSLE)} & \textbf{state (F1)} & \textbf{$\alpha$ Ens. (W1)} & \textbf{$K$ Ens. (W1)} \\
\midrule
1 & 1280 & 0 & 0.005 & 0.082 & 0.044 & 1 & 0.146 & 0.021 \\
2 & 8983 & 0 & 0.055 & 0.172 & 0.021 & 0.994 & 0.119 & 0.027 \\
3 & 1348 & 1.045 & 0.642 & 0.125 & 0.042 & 0.868 & 0.141 & 0.116 \\
4 & 1192 & 1.485 & 0.689 & 0.159 & 0.032 & 0.650 & 0.142 & 0.012 \\
5 & 6485 & 1.059 & 0.303 & 0.206 & 0.064 & 0.993 & 0.149 & 0.041 \\
6 & 1172 & 1.140 & 0.484 & 0.206 & 0.015 & 0.995 & 0.065 & 0.018 \\
7 & 1222 & 1.761 & 0.048 & 0.256 & 0.047 & 1 & 0.069 & 0.021 \\
8 & 1524 & 0 & 0.997 & 0.170 & 0.581 & 0.972 & 0.129 & 2.403 \\
9 & 1857 & 2.459 & 0.040 & 0.877 & 0.153 & 0.233 & 0.784 & 0.090 \\
\midrule
\textbf{Avg.} & \textbf{2785} & \textbf{0.722} & \textbf{0.254} & \textbf{0.231} & \textbf{0.080} & \textbf{0.914} & \textbf{0.194} & \textbf{0.305} \\
\bottomrule
\end{tabularx}
\end{table}

\section{Conclusions}
\label{sec:conclusions}

AnomalousNet is an integrated pipeline that combines a particle tracking algorithm, an Attention U-Net, and a change-point detection method to infer anomalous diffusion parameters directly from video data. The inherent complexity of this task is underscored by the fact that only three entries addressed the video track in the 2nd AnDi Challenge, with AnomalousNet emerging as one of the few successful solutions \cite{munoz-gil2023objective}.\medskip

Despite the challenges posed by noisy, high-dimensional video data and short, overlapping trajectories, the system accurately estimated the anomalous diffusion exponent, diffusion coefficient, and state transitions. This performance demonstrates the potential of blending advanced deep learning techniques with robust signal processing to tackle complex physical phenomena.\medskip

Overall, while further improvements, especially in the change-point detection module, are possible, this work establishes a promising framework for analyzing anomalous diffusion in video datasets and provides a modular system that can be refined in individual components without the need to build an entirely new system from scratch.\medskip

\section*{Acknowledgements}

J.A.C. is supported by the European Union - NextGenerationEU, ANDHI project CPP2021-008994 and PID2021-124618NB-C21, by MCIN/AEI/10.13039/501100011033 and by “ERDF A way of making Europe”, from the European Union.

\clearpage

\section*{References}
\bibliographystyle{unsrt}  
\bibliography{references}   

\end{document}